\documentclass[conference]{IEEEtran}
\IEEEoverridecommandlockouts                             

\usepackage{booktabs}
\usepackage{graphicx}
\usepackage{color}
 \usepackage{url} 
 \usepackage{hyperref}
  \usepackage{comment} 
 
 \usepackage{cite}
\usepackage{amsmath,amssymb,amsfonts}

\usepackage{algorithmic}
\usepackage{graphicx}
\usepackage{textcomp}
\usepackage{xcolor}
\usepackage{seqsplit}
\usepackage{tikz}
\usepackage{tabularx}
\usepackage{array}
\usepackage{listings}
\usepackage{adjustbox}
\usepackage{multirow}
\newcolumntype{M}[1]{>{\centering\arraybackslash}m{#1}}
\usetikzlibrary{shapes.geometric,shapes.symbols,fit,positioning,shadows, arrows.meta, backgrounds}

\def\BibTeX{{\rm B\kern-.05em{\sc i\kern-.025em b}\kern-.08em
    T\kern-.1667em\lower.7ex\hbox{E}\kern-.125emX}}
\begin{document}
 \makeatletter
\newcommand{\linebreakand}{%
\end{@IEEEauthorhalign}
\hfill\mbox{}\par
\mbox{}\hfill\begin{@IEEEauthorhalign}
}
\makeatother
\title{Customizing Visual-Language Foundation Models for Multi-Modal Anomaly Detection and Reasoning 
}
\author{
	\IEEEauthorblockN{Xiaohao Xu\IEEEauthorrefmark{1}\IEEEauthorrefmark{2}, Yunkang Cao\IEEEauthorrefmark{1}\IEEEauthorrefmark{3}, Huaxin Zhang\IEEEauthorrefmark{3}, Nong Sang\IEEEauthorrefmark{3}, Xiaonan Huang\IEEEauthorrefmark{2}}
	\linebreakand
	\IEEEauthorblockA{
    	\IEEEauthorrefmark{1} Equal Contribution
	\quad
	\IEEEauthorrefmark{2} Robotics Department, University of Michigan-Ann Arbor, Ann Arbor, MI, 48105, USA.
	\\
	\IEEEauthorrefmark{3} School of Mechanical Science and Engineering, Huazhong University of Science and Technology, Wuhan, 430074, China. 
    \\
    \{xiaohaox, xiaonanh\}@umich.edu, \{cyk\_hust, zhanghuaxin, nsang\}@hust.edu.cn
}
}

\maketitle
\begin{abstract}
Anomaly detection is vital in various industrial scenarios, including the identification of unusual patterns in production lines and the detection of manufacturing defects for quality control. Existing techniques tend to be specialized in individual scenarios and lack generalization capacities. In this study, our objective is to develop a generic anomaly detection model that can be applied in multiple scenarios. To achieve this, we custom-build generic visual language foundation models that possess extensive knowledge and robust reasoning abilities as anomaly detectors and reasoners. Specifically, we introduce a multi-modal prompting strategy that incorporates domain knowledge from experts as conditions to guide the models. Our approach considers diverse prompt types, including task descriptions, class context, normality rules, and reference images. In addition, we unify the input representation of multi-modality into a 2D image format, enabling multi-modal anomaly detection and reasoning. Our preliminary studies demonstrate that combining visual and language prompts as conditions for customizing the models enhances anomaly detection performance. The customized models showcase the ability to detect anomalies across different data modalities such as images, point clouds, and videos. Qualitative case studies further highlight the anomaly detection and reasoning capabilities, particularly for multi-object scenes and temporal data.   Our code is publicly available at \href{https://github.com/Xiaohao-Xu/Customizable-VLM}{\url{https://github.com/Xiaohao-Xu/Customizable-VLM}}.\footnote{More insights of customized foundation models for broader anomaly detection settings are available at Github repo: \href{GPT4V-for-Generic-Anomaly-Detection}{\url{https://github.com/caoyunkang/GPT4V-for-Generic-Anomaly-Detection}}.}
\end{abstract}
 

\section{INTRODUCTION}






Traditional model development paradigms for industry often results in solutions that are specialized yet plagued by inefficiency, limited scalability, and a lack of flexibility when adapting to new manufacturing settings. 
However, recent advancements in foundation models, such as the large-scale visual-language model ChatGPT~\cite{gpt4v}, have shown the potential to revolutionize this industrial landscape. 
These models offer impressive abilities in understanding and generating human-like linguistic texts, thereby enabling more interpretable and natural interfaces for human-machine interaction. Moreover, the rapid progress in visual-language models has demonstrated their robust reasoning capabilities~\cite{clip}, making them well-suited for understanding industrial settings and rules at a higher level. This progress opens exciting avenues for unifying diverse industrial applications.



This study focuses on anomaly detection, which aims to identify data patterns or behaviors that largely deviate from established norms~\cite{MVTec-AD}. Anomaly detection plays a vital role in various industrial applications, particularly in automatic defect inspection and fault diagnosis. Recognizing unusual patterns helps prevent failures and ensures quality control. Traditional approaches to anomaly detection have typically been designed for specific categories~\cite{MVTec-AD}. However,  a generic model for anomaly detection has the potential to both unify anomaly detection across different contexts and enable high-level reasoning and explanation of these anomalies. 

\begin{figure}[t!]
    \centering
\includegraphics[width=0.48\textwidth]{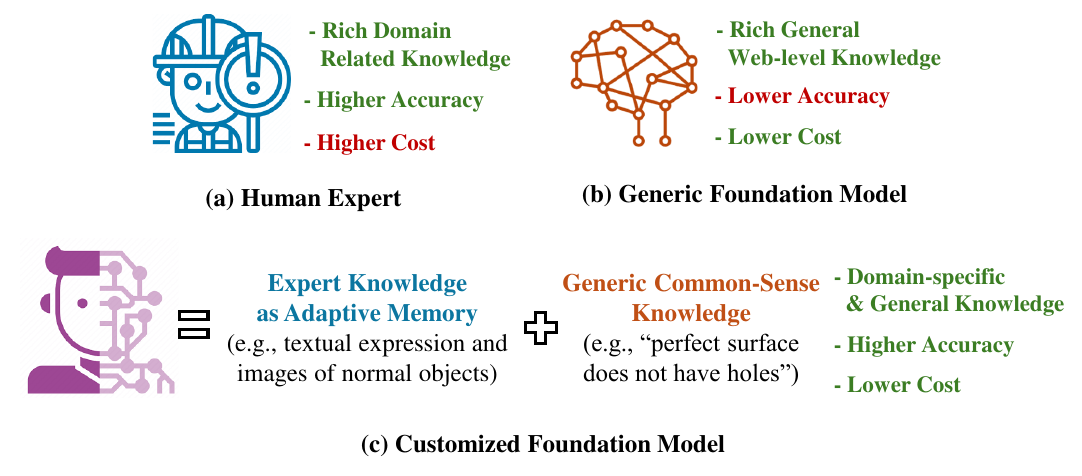}\vspace{-3mm}
    \caption{\textbf{Human-model collaboration for customizing Visual-Language Foundation Models in anomaly detection and reasoning.} We incorporate expert domain knowledge as external memory and instructions, enabling the model to leverage both specialized expertise and general reasoning capabilities. This collaborative prompting approach enhances the model’s ability to detect anomalies and provide context-rich explanations.} 
    \label{fig:teaser}\vspace{-5mm}
\end{figure}

\begin{figure*}[t!]
    \centering\setlength{\abovecaptionskip}{0.1cm}
    \includegraphics[width=\textwidth]{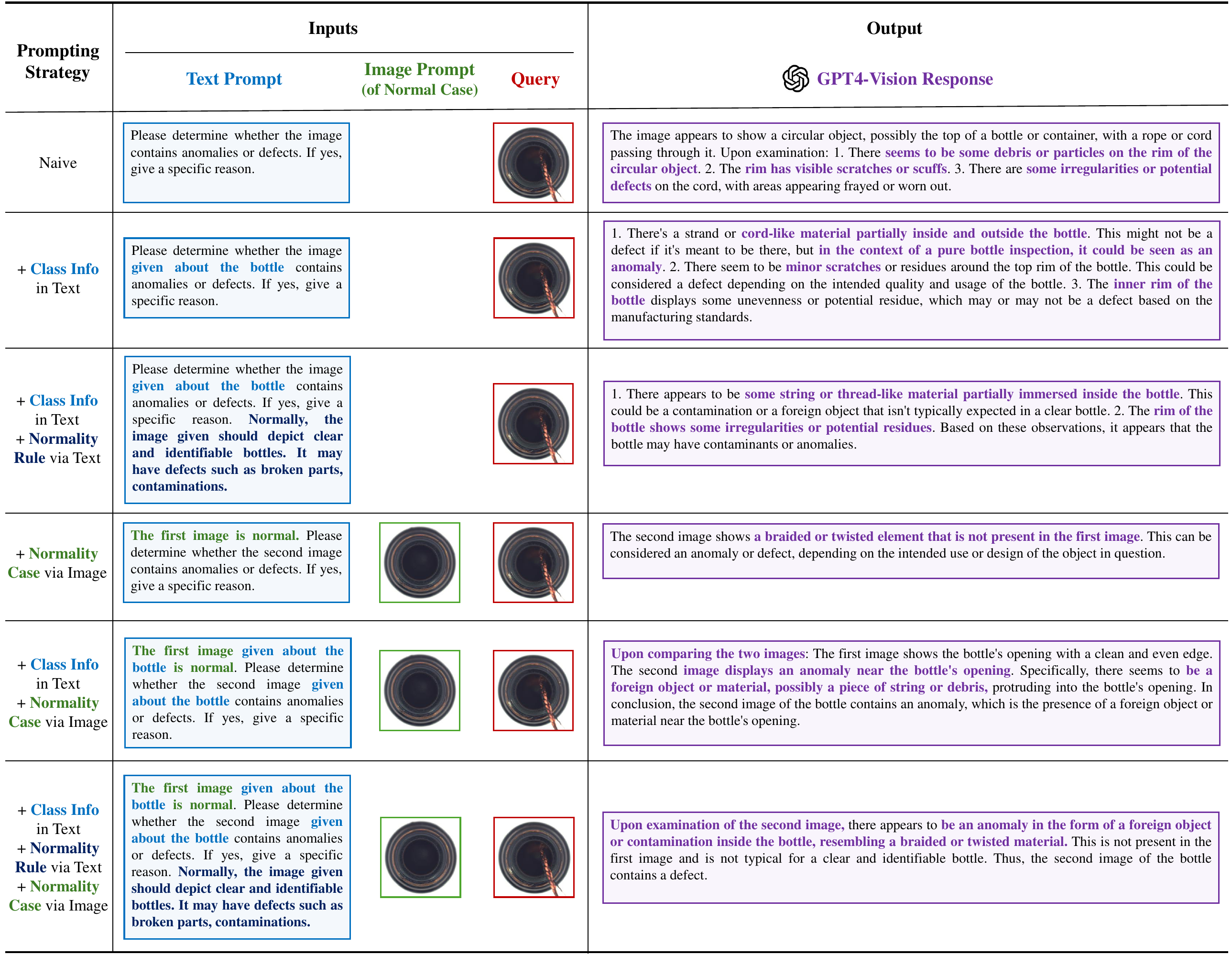}
\vspace{-5mm}
\caption{\noindent\textbf{Prompt taxonomy for prompting engineering.} \textit{1) Naive}: framing the task as a simple sentence instruction; \textit{2) Class Info}: adding target class information for context; \textit{3) Normality Rule}: defining normality and abnormality through textual rules; and \textit{4) Visual reference}: normal images for reference.}
    \label{fig:prompt}\vspace{-4mm}
\end{figure*}

Although generic foundation models possess extensive knowledge, they may not fully meet the requirements of diverse industrial anomaly detection due to the varied contextual nature of anomalies across different industries. Anomalies can vary significantly depending on the specific industrial domain. For example, in manufacturing, an incorrect order of two parts might be considered a logical anomaly~\cite{MVTec-LOCO}. To effectively harness the capabilities of generic foundation models for anomaly detection, it is crucial to integrate domain-specific knowledge, such as rules for anomaly identification and target object distribution, into these models.

To achieve this, our objective is to extract knowledge and contextual information from industry experts and integrate this expertise into the foundation model. As shown in Fig.~\ref{fig:teaser}, industry experts have domain-specific knowledge that enables them to accurately identify defects by recognizing normal and abnormal patterns. Our approach involves incorporating their expertise as external adaptive memory prompts, such as textual or visual representations of normal objects. These prompts serve as priors for the generic foundation model, allowing it to reason about anomalies within a specific context. By combining domain-specific prior knowledge with the foundation model's general intelligence, we create customized industrial foundation models. These customized models are expected to achieve higher accuracy than the generic foundation model by incorporating human expertise through the prompts. The objective of this study is to evaluate the anomaly detection and reasoning capabilities of existing large-scale visual-language models, particularly focusing on exploring prompt engineering to leverage their comprehensive knowledge base effectively.

Our main contributions are summarized as:

\textbf{1}) Our work is among the first to employ visual-language models for multi-modal anomaly detection and reasoning. 

\textbf{2}) We present a prompting approach using task instructions, class context, normality rules, and reference examples, increasing detection accuracy and interpretability. 

\textbf{3}) Our case studies show that tailoring VLMs with domain-specific prompts enables promising anomaly detection across diverse modalities (including 2D images, 3D point cloud, and video), widening industrial applicability.

\section{Methodology}
In this section, we outline the framework for transforming a generic foundation model into a customized foundation model for anomaly detection and reasoning.

\subsection{Overview}
The process of customizing a foundation model for anomaly detection involves using a generic visual language foundation model $\Phi$. We provide a query sample $\mathbf{q}$ along with a sophisticated instruction, known as a prompt $\mathbf{p}$, to the model. The prompt can be images or texts. The model then predicts the binary detection result $\mathbf{c} \in \{0, 1\}$ and a textual reasoning response $\mathbf{r}$.
This process is given by: $
\mathbf{c}, \mathbf{r} = \Phi(\mathbf{q}, \mathbf{p})$.

\subsection{Unified Pre-processing for Multi-modal Inputs}
For industrial applications, the query sample $\mathbf{q}$ can vary widely in modality, such as RGB images, point clouds, and time series data. Traditional anomaly detection models have typically been modality-specific, developed separately for each type of data. In contrast, we propose a unified pre-processing operation that converts all data into a standardized 2D image format. This standardization facilitates the application of large-scale visual-language models across different modalities by ensuring consistent input structure. Specifically, point clouds are projected onto a 2D plane to obtain rendered depth images; time-series data are visualized and plotted as figures. 

\subsection{Multi-modal Prompt Engineering}

\noindent\textbf{Prompt taxonomy.} To customize a generic foundation model for anomaly detection and reasoning, we introduce domain-specific contextual knowledge using four primary types of prompts $\mathbf{p}$ (see Fig. \ref{fig:prompt}). These prompts infuse the model with a deeper understanding of the task and domain-specific nuances.

\noindent\textit{1) Task instruction}. Clear task information prompts the generic foundation model for effective anomaly detection.

\noindent\textit{2) Class context}. Explicit class context information, represented by a class token [CLS], enhances the model's recognition of the target domain. We use object type names like `bottle' and `candle' as the [CLS] token.

\noindent\textit{3) Normality criteria}. Language-form rules define explicit normal standards and describe abnormal objects and patterns based on human expertise. For example, `The given image should depict a clean and well-structured PCB with clear traces, soldered components, and distinct labels'.

\noindent\textit{4) Reference normal image}. A normal reference image of the target class aids in identifying abnormal distributions by comparing them with the reference. It helps the model understand normality better, as describing anomaly regions concisely through text alone can be challenging.

\begin{table*}[h!]
\centering \setlength{\abovecaptionskip}{0.1cm} 
\caption{ Anomaly Detection Performance on MVTec-AD dataset.}
\label{tab:object_metrics}

\resizebox{0.8\textwidth}{!}{
\begin{tabular}{l|ccc|ccc|ccc|ccc}
\toprule[1.5pt]
&  \multicolumn{6}{c|}{{Commercial Foundation Models}} & \multicolumn{6}{c}{{Open-sourced Foundation Models}}\\ \midrule
Base Model & \multicolumn{3}{c|}{{Gemini Vision  Pro 1.0}} & \multicolumn{3}{c|}{{GPT4-V(ision)}} & \multicolumn{3}{c|}{{InternVL2-8B}} & \multicolumn{3}{c}{{Qwen2VL-7B}} \\
\midrule
{Category} & {ACC} & {AUROC} & {AUPR} & {ACC} & {AUROC} & {AUPR} & {ACC} & {AUROC} & {AUPR} & {ACC} & {AUROC} & {AUPR} \\
\midrule
carpet & 0.891 & 0.854 & 0.646 & \textbf{0.957} & \textbf{0.935} & \textbf{0.852} &0.744 &0.489 &0.239 &0.897 &0.933 &0.700\\
grid & 0.820 & 0.571 & 0.323  & {0.938} & {0.914} & {0.850} &0.731 &0.500 &0.269 & \textbf{0.987} &\textbf{0.991} &\textbf{0.955}\\
leather  & 0.903 & 0.812 & 0.722 & \textbf{0.984} & \textbf{0.979} & \textbf{0.947} &0.742 &0.500 &0.258 & 0.952 &0.967 &0.842\\
tile & 0.829 & 0.734 & 0.554  & \textbf{0.957} & \textbf{0.961} & \textbf{0.870}  &0.709 &0.494 &0.282  &0.915 &0.931 &0.765\\
wood & 0.835 & 0.658 & 0.480 & \textbf{0.987} & \textbf{0.974} & \textbf{0.960}   &0.367 &0.583 &0.275  &0.962 &0.957 &0.865\\
bottle & 0.867 & 0.742 & 0.575  & \textbf{0.964} & \textbf{0.959} & \textbf{0.872} &0.783 &0.601 &0.337 &0.783 &0.857 &0.526\\
cable & 0.627 & 0.517 & 0.408 & \textbf{0.834} & \textbf{0.814} & \textbf{0.672} &0.613 &0.500 &0.387 &0.667 &0.728 &0.537\\
capsule & \textbf{0.803} & 0.658 & 0.288 & 0.674 & \textbf{0.803} & \textbf{0.348} &0.788 &0.477 &0.174 &0.447 &0.665 &0.240\\
hazelnut & 0.655 & 0.541 & 0.396 & {0.845} & {0.787} & {0.730}  &0.409 &0.536 &0.381 &\textbf{1.000} &\textbf{1.000} &\textbf{1.000}\\
metal nut& 0.774 & 0.565 & 0.229  & \textbf{0.890} & {0.783} & \textbf{0.550} &0.809 &0.500 &0.191 &0.696 &\textbf{0.812} &0.386\\
pill & \textbf{0.856} & \textbf{0.648} & \textbf{0.297} & 0.820 & 0.486 & 0.156 &0.844 &0.500 &0.156 &0.677 &0.809 &0.325\\
screw & \textbf{0.745} & 0.627 & 0.307 & 0.669 & \textbf{0.769} & \textbf{0.430} &0.744 &0.500 &0.256 &0.662 &0.725 &0.397\\
toothbrush & 0.786 &\textbf{ 0.675} & 0.464 & \textbf{0.810} & 0.667 & \textbf{0.524} &0.619 &0.558 &0.315 &0.714 &0.600 &0.357\\
transistor  & 0.400 & 0.500 & 0.600 & {0.750} & {0.700} & {0.715} &0.400 &0.500 &0.600 &\textbf{0.840} &\textbf{0.817} &\textbf{0.809}\\
zipper & 0.656 & {0.736} & {0.349} & {0.689} & 0.723 & 0.347 &\textbf{0.788} &0.500 &0.212 &0.768 &\textbf{0.853} &\textbf{0.478}\\
\midrule
average & {0.760} & {0.625} & {0.370}  & \textbf{0.842} & {0.829} & \textbf{0.593} &0.691 &0.520 &0.281 &0.780 &\textbf{0.836} &0.542\\
\bottomrule[1.5pt]
\end{tabular}}\label{tab:benchmark}\vspace{-4mm}
\end{table*}

\begin{figure}[t!]
    \centering \setlength{\abovecaptionskip}{0.0cm} 
    \setlength{\abovecaptionskip}{0.0cm}\includegraphics[width=0.48\textwidth]{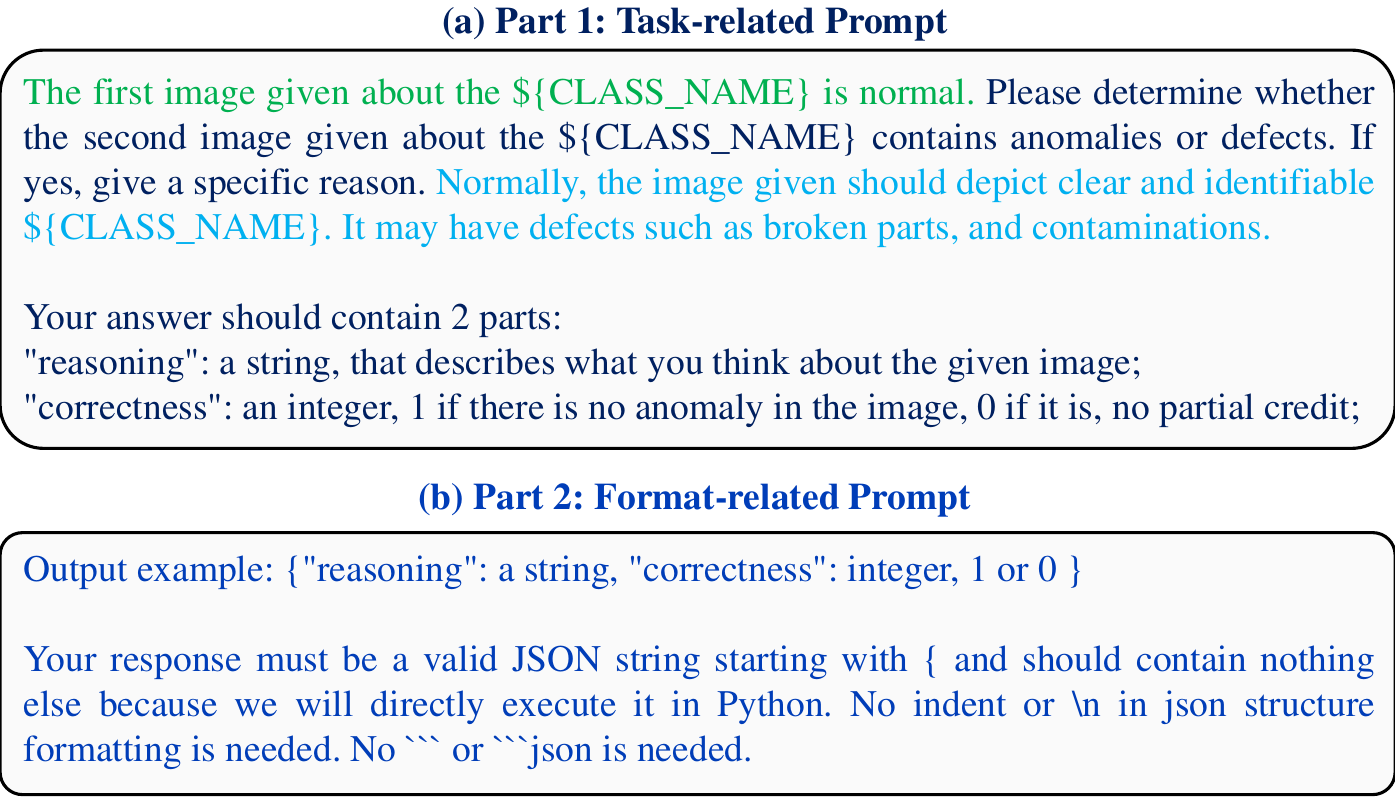}
 \caption{{{\textbf{Prompting template for anomaly detection and reasoning.}}}}
    \label{fig:prompting_template}  \vspace{-4mm}
\end{figure}

\begin{figure}[t!]
    \centering
\setlength{\abovecaptionskip}{0.0cm}   \includegraphics[width=0.5\textwidth]{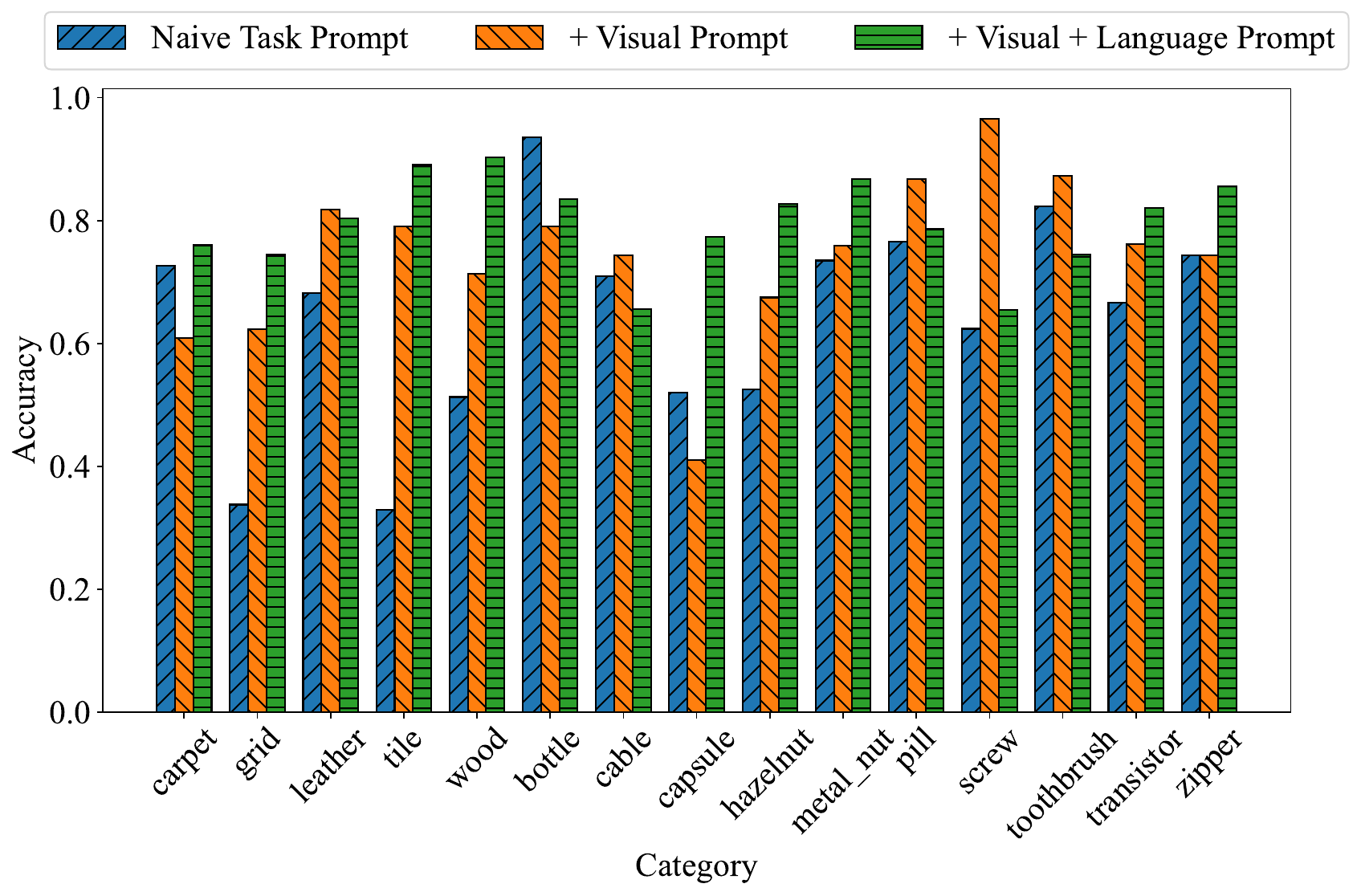}
 \caption{{\textbf{Ablation study of the prompting strategy using Gemini Pro Vision 1.0}. The visual prompts indicate the reference normal image, while the language prompts represent the normality rules.}}
    \label{fig:ablation_quantitative}\vspace{-4mm}
\end{figure}

\begin{figure*}[hbt!]
    \centering
    \includegraphics[width=\textwidth]{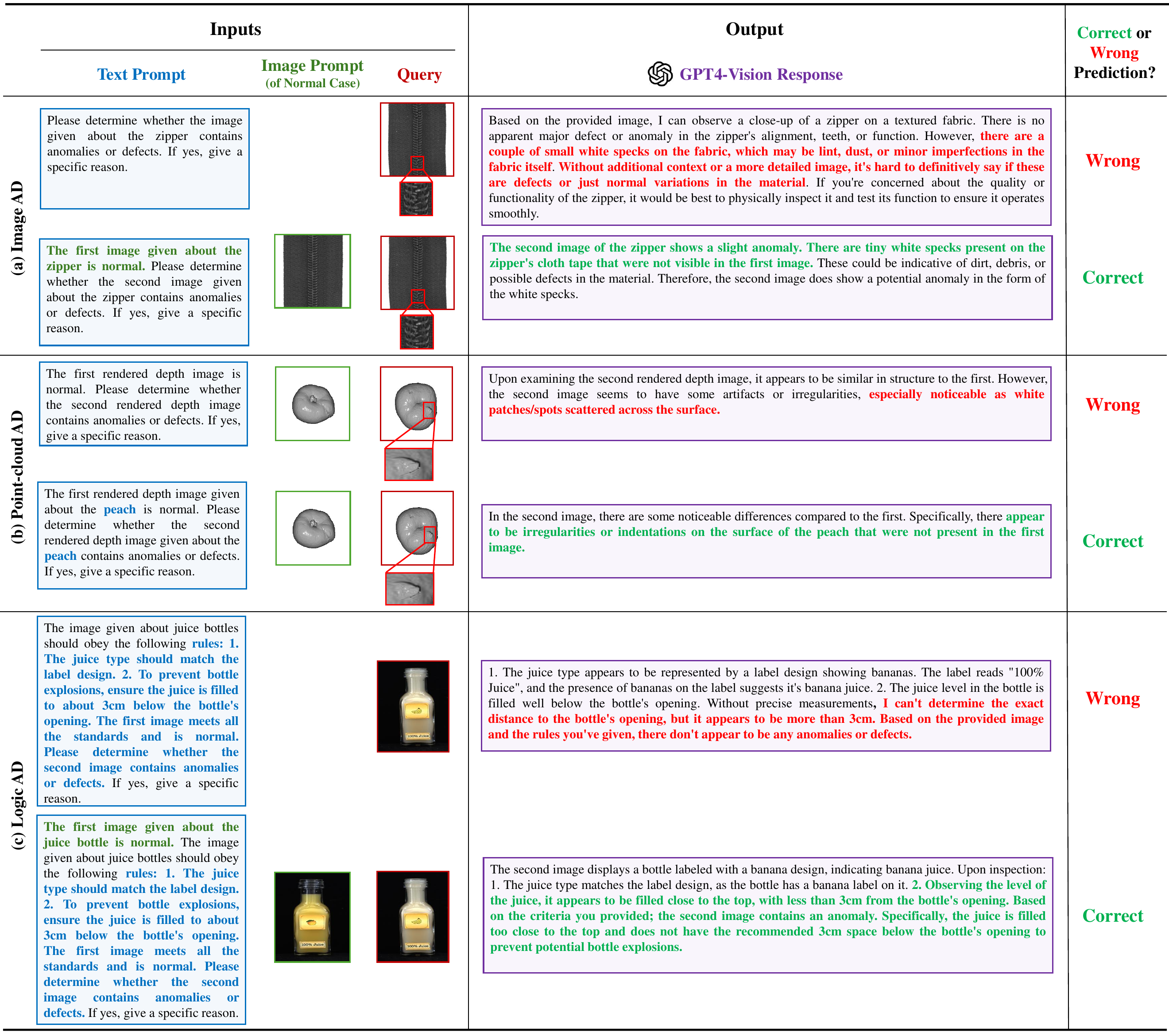}\vspace{-2mm}
 \caption{{\textbf{Effect of using multi-modal prompts}, including language-based and visual cues, as conditions to customize foundation models.}}
    \label{fig:ablation}\vspace{-3mm}
\end{figure*}

\noindent\textbf{Prompt template for anomaly detection and reasoning.} As shown in Fig.~\ref{fig:prompting_template}, we use a prompt that combines task-related and format-related information to control the output format of the foundation models for further quantitative analysis. The task-related prompt defines the anomaly detection task, while the format-related prompt instructs the model to output the anomaly detection in binary form (0/1) and provide the reasoning response for the prediction via Python string. 

\section{Experiment}

\subsection{Quantitative Results of Anomaly Detection}


\begin{figure*}[t!]
    \centering\setlength{\abovecaptionskip}{0.2cm}
    \includegraphics[width=\textwidth]{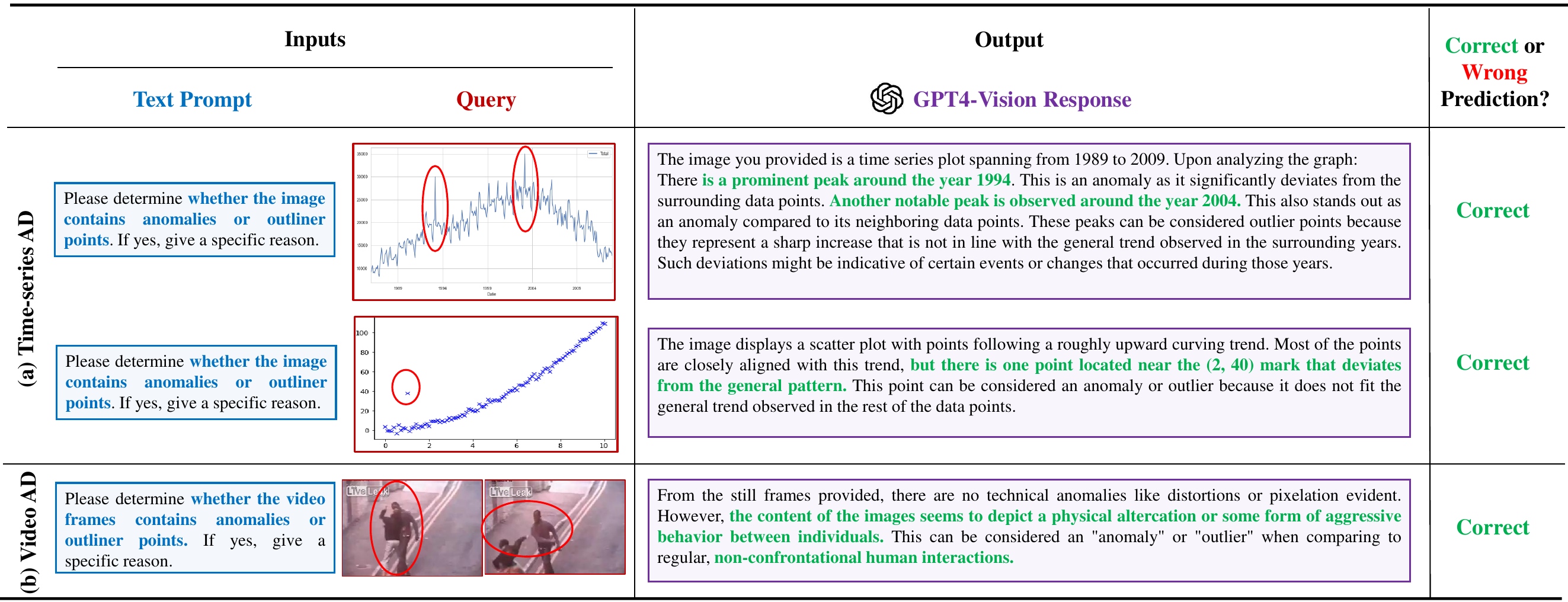}\vspace{-2mm}
    \caption{{\textbf{Temporal anomaly detection capability} encompassing (\textbf{a}) time-series signals and (\textbf{b}) video-level sequences.}
}
    \label{fig:beyond}\vspace{-2mm}
\end{figure*}

\noindent\textbf{Benchmark results.}  
We evaluate a mix of commercial and open-source vision-language models (VLMs) on the MVTec-AD~\cite{MVTec-AD} dataset using our proposed prompting strategy. We choose the first normal image from the training set as the reference image when visual prompt is enabled.
 Commercial models include Gemini Pro Vision 1.0~\cite{team2023gemini} and GPT4-V~\cite{gpt4v}, while open-source models evaluated are Qwen2VL~\cite{Qwen2VL} and InternVL2~\cite{internvl2}. Table~\ref{tab:benchmark} details the anomaly detection performance across key metrics: Accuracy (ACC), Area Under the Receiver Operating Characteristic Curve (AUROC), and Area Under the Precision-Recall Curve (AUPR).

Our benchmarking reveals several key insights. \textbf{1}) GPT4-V consistently outperforms other models in ACC and AUPR metrics for most object categories, achieving an average accuracy of 84\% without any additional training or prompt tuning. This demonstrates its robustness in handling diverse anomaly detection tasks with minimal configuration. \textbf{2}) The recently released open-source model Qwen2VL-7B (August 2024) outperforms the commercial models, \textit{i.e.}, Gemini and GPT4-V, in AUROC, highlighting the rapid progress and competitiveness of open-source foundation models. \textbf{3}) While GPT4-V excels in ACC and AUPR, Qwen2VL-7B shows strengths in AUROC, suggesting complementary advantages between the models, with their differences reflecting distinct optimization strategies and model architectures.  \textbf{4}) At the category level, the models display domain-specific strengths. For example, GPT4-V achieves outstanding results in categories with well-defined visual features such as `carpet' while Qwen2VL-7B shows greater resilience to subtle anomalies in less structured categories like `hazelnut'.


\noindent\textbf{Ablation study of prompting strategy.}  
In Fig.~\ref{fig:ablation_quantitative}, we analyze the influence of various prompting strategies on the MVTec-AD dataset~\cite{MVTec-AD} using the Gemini Vision Pro 1.0 base model. Visual prompts utilize the reference normal image, while language prompts capture the normality rules. The results show significant improvements across most sub-categories when incorporating cues from both the reference normal image and normality rules. In addition, our findings indicate that sub-categories such as `grid', `tile', and `wood' exhibit the lowest accuracy when using the naive prompt. However, we observe that incorporating additional visual and language prompts can significantly enhance their performance.

\subsection{Quantitative Case Studies of Anomaly Reasoning}

\noindent\textbf{Effect of diverse prompting strategies.}  We assess the impact of various prompting strategies on anomaly detection performance utilizing the MVTec-AD dataset~\cite{MVTec-AD}, known for its diverse range of industrial anomaly images. As depicted in Fig.~\ref{fig:prompt}, employing sophisticated prompts and combining textual normality rules with image references greatly enhances the reasoning for anomalies and reduces ambiguity.
For instance, using a naive prompt (first column in Fig.~\ref{fig:prompt}) leads to imprecise anomaly hypotheses, with uncertain responses (\textit{e.g.}, `seems to be some debris') due to the model's limited knowledge of the normality distribution.
In contrast, incorporating external knowledge and contextual prompts improves the accuracy of the model. It can accurately describe the anomaly object (\textit{e.g.}, `a braided or twisted material') by comparing it with the reference normal image (\textit{e.g.}, `This is not present in the first image and is not typical for a clear and identifiable bottle'). This feature allows users to enhance the model's performance by providing additional information without training the model.

\begin{figure*}[ht!]
    \centering
    \includegraphics[width=\textwidth]{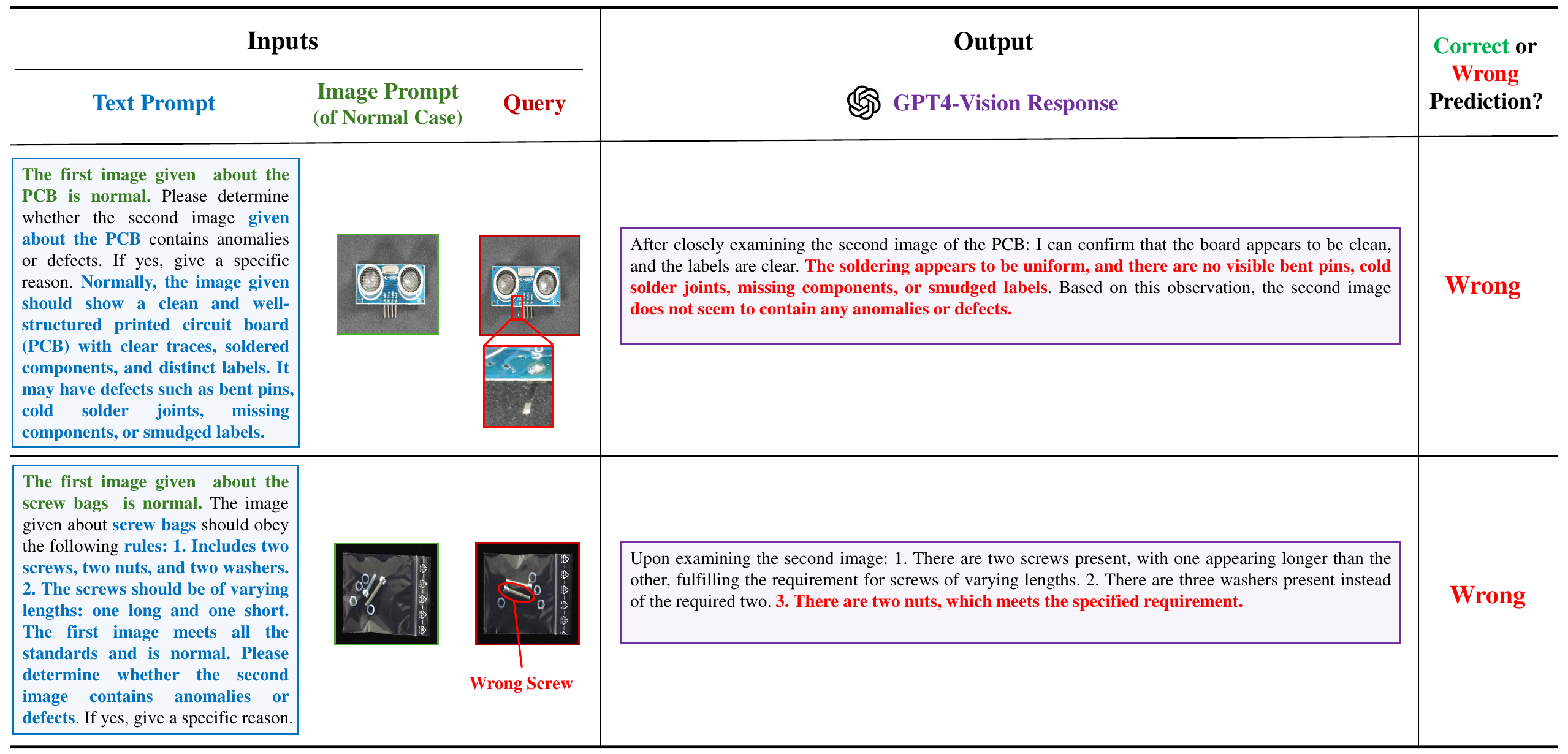}\vspace{-3mm}
\caption{\textbf{Failure cases.} Customized foundation models modulated by prompts fail to detect small defects (\textbf{Top}) and reason over complex scenes (\textbf{Bottom}).}
    \label{fig:failure}\vspace{-3mm}
\end{figure*}

\noindent\textbf{Multi-modal (image and 3D) anomaly detection ability.}
The versatility of adapted GPT4-V in handling diverse data modalities, including images~\cite{MVTec-AD} and point clouds~\cite{MVTec-3D}, is showcased in Fig.~\ref{fig:ablation}. The model exhibits its ability to incorporate external knowledge, such as language-based rules defining normality and abnormality, as well as reference images of normal target objects. This integration leads to improved anomaly detection results and accurate responses.
Specifically, Fig.~\ref{fig:ablation} (a) demonstrates that introducing an additional reference normal image markedly improves the model's ability to detect anomalies on the zipper in the image, effectively filtering out irrelevant image noise (\textit{e.g.}, dust). Furthermore, Fig.~\ref{fig:ablation} (b) showcases that the adapted foundation model accurately identifies the presence of a small protrusion or bump on the right part of the peach's surface, rendered from the 3D geometry. These findings highlight the capability of current generic models to address multi-modal anomaly detection.

\noindent\textbf{Global logical anomaly detection capacity.} Fig.~\ref{fig:ablation} (c) demonstrates the anomaly detection capability of GPT4-V, using a specific case involving a bottle with an excessive amount of juice from the MVTec-LOCO~\cite{MVTec-LOCO} dataset. The model exhibits a comprehensive understanding of global semantics, enabling it to identify overarching abnormal patterns or behaviors. One notable feature of the adapted GPT4-V model is its ability to autonomously reason about complex normal standards and provide rationales for detected anomalies (\textit{e.g.}, `Observing the level
of the juice, it appears to be filled close to the top'). However, without sufficient reference information from the prompt, the model may struggle to reason about the anomaly accurately. This emphasizes the importance of providing detailed reference information for reliable model prediction.

\noindent\textbf{Temporal anomaly detection ability.}
Anomaly detection for time series data~\cite{shen2020timeseries} is essential in various industrial applications, especially in monitoring equipment performance during manufacturing processes. 
In Fig. \ref{fig:beyond} (a) (top case), the model effectively identifies abnormal peaks, such as a prominent peak around the year 1994. Moreover, it provides a detailed explanation, correctly attributing these peaks to outlier points that represent a significant increase inconsistent with the general trend observed in the surrounding years.
In Fig. \ref{fig:beyond} (a) (bottom case), GPT4-V successfully detects an anomaly at an outlier point and astonishingly provides the precise coordinates of that point. For example, it outputs a reason like `There is one point located near the (2, 40) point mark that deviates from the general pattern'.
Moreover, Fig. \ref{fig:beyond} (b) depicts a scenario where two people collide from the UCF-Crime dataset~\cite{sultani2018real}. Even with a naive task instruction (\textit{e.g.}, `determine whether the video frames contain anomalies or outlier points'), GPT4-V successfully detects the physical altercation between the two individuals. Moreover, it accurately accounts for the anomaly for non-confrontational human interactions.

\noindent\textbf{Failure cases.}
In Fig. \ref{fig:failure}, we highlight the intrinsic limitations of current generic foundation models, which impair their performance when prompted for anomaly detection. Specifically, the model cannot recognize a bent pin on an ultrasonic sensor due to its inability to discern fine details (Fig. \ref{fig:failure} (top case)). Similarly, in the bottom case of Fig. \ref{fig:failure}, it overlooks an anomaly where a screw is missing within a plastic pocket. These instances highlight the limitations of current models in understanding the relationships among visual elements.

\section{Conclusion and Future Works}

\noindent\textbf{Conclusion.} We proposed to customize generic visual-language foundation models for anomaly detection and reasoning using multi-modal prompting strategies. The experiments demonstrated the effectiveness of combining visual and language prompts. The adapted models demonstrated notable success across benchmark datasets and showcase the ability to detect and reason anomalies in diverse modalities and scenarios. Despite these promising results, challenges persist, especially in capturing fine-grained details, highlighting the need for continuous refinement to unlock the full potential of foundation models in real-world anomaly detection tasks.

\noindent\textbf{Future works.}
Customizing foundation models using proper prompts shows great promise for anomaly detection across domains. To further enhance anomaly detection performance, we can employ multi-round conversations with foundation models for iterative learning and investigate the integration of additional sensor readings or metadata.


\bibliographystyle{plain} 
\bibliography{main} 

\end{document}